\documentclass[12pt]{article}
\usepackage{amsfonts}
\usepackage{amsmath}
\usepackage{mathrsfs}
\usepackage{color}
\usepackage{tikz}
\usepackage{authblk}
\usepackage{mathrsfs,amscd,amssymb,amsthm,amsmath,bm,graphicx,psfrag,subfigure,url}

\usepackage{algorithmic}
\usepackage{algorithm}

\usetikzlibrary{positioning, shapes.geometric}

\allowdisplaybreaks
\usepackage{indentfirst}

\def \[{\begin{equation}}
\def \]{\end{equation}}

\newtheorem{thm}{Theorem}[section]
\newtheorem{Proposition}{Proposition}
\newtheorem{defi}{Definition}

\newtheorem{cor}[thm]{Corollary}
\newtheorem{exa}[thm]{Example}

\newtheorem{ass}{Assuption}

\begin{document}
\title{\bf An upper bound of the mutation probability in the genetic algorithm for general 0-1 knapsack problem}
\author{Yang Yang\thanks{Email: zhugemutian@outlook.com}}
\affil{School of Mathematical Sciences, Xiamen
University, Xiamen 361005, P.R. China}
\date{}
\maketitle

\begin{abstract}
As an important part of genetic algorithms (GAs), mutation operators is widely used in evolutionary algorithms to solve $\mathcal{NP}$-hard problems because it can increase the population diversity of individual. Due to limitations in mathematical tools, the mutation probability of the mutation operator is primarily empirically set in practical applications.

In this paper, we propose a novel reduction method for the 0-1 knapsack problem(0-1 KP) and an improved mutation operator (IMO) based on the assumption $\mathcal{NP}\neq\mathcal{P}$, along with the utilization of linear relaxation techniques and a recent result by Dey et al. (Math. Prog., pp 569-587, 2022). We employ this method to calculate an upper bound of the mutation probability in general instances of the 0-1 KP, and construct an instance where the mutation probability does not tend towards 0 as the problem size increases. Finally, we prove that the probability of the IMO hitting the optimal solution within only a single iteration in large-scale instances is superior to that of the traditional mutation operator.
\end{abstract}

\vspace{2mm} \noindent{\bf Keywords}: genetic algorithm; mutation operator; mutation probability; upper bound; reduction method
\vspace{2mm}

\setcounter{section}{0}

\section{Introduction}

\begin{defi}\cite{MT1990, KPP2004, Cacchianietal2022a, Cacchianietal2022b}
Given a set $N$ consisting of $n$ items, where the $j$-th item has a profit and a weight, and denoted by $p_j$ and $w_j$, respectively. The goal of the 0-1 knapsack problem(0-1 KP) is to find an optimal subset of the set $N$ such that its total profit is maximized without exceeding the knapsack capacity $C$. The problem can be formulated as follows.
\begin{align}
    \max f(\textbf{X})=\sum\limits_{j=1}^n x_jp_j,\label{KP-model-obj}
\end{align}
s.t.
\begin{align}
    g(\textbf{X})=\sum\limits_{j=1}^n x_jw_j\le C,\label{KP-model-con-1}\\
    x_j\in\{0,1\},\label{KP-model-con-2}
\end{align}
\end{defi}
where $x_j=1$ represents the $j$-th item is packed in the knapsack while $x_j=0$ indicates not. The objective function (\ref{KP-model-obj}) of the model aims to maximize the total profit of the selected items. Inequality (\ref{KP-model-con-1}) is a capacity constraint that ensures the total weight of selected items do not exceed the knapsack capacity $C$. The decision variables $x_j$ are binary, as indicated in (\ref{KP-model-con-2}).

A substantial number of commercial decision-making and industrial manufacturing problems can be formulated as the 0-1 KP, such as budget control and cargo loading. This has resulted in a significant focus on developing efficient algorithms to tackle these problems. However, the finding for a polynomial-time algorithm to exactly solve the 0-1 KP, which is widely recognized as an $\mathcal{NP}$-hard problem\cite{Cook1971, Karp1972, GJ1979}, poses a substantial challenge. Existing algorithms like dynamic programming \cite{Bellman1957, Toth1980} can only solve the problem exactly within a pseudo-polynomial time complexity of $O(nC)$, and even the fastest exact algorithms require $O(2^{n/2})$ \cite{HS1974, SS1981}. Additionally, a significant body of research indicates that unless one intends to demonstrate $\mathcal{NP} = \mathcal{P}$, we can assume $\mathcal{NP} \neq \mathcal{P}$ \cite{SBR2002, DS2005}. If $\mathcal{NP} \neq \mathcal{P}$, for a general instance of the 0-1 KP, there are items whose selection in the optimal solution cannot be determined without exhaustively enumerating all feasible solutions.

Coincidentally, in the 1960s, Holland proposed a genetic algorithm(GA) \cite{Holland1975} that simulates biological evolution, consisting of selection, crossover, and mutation operators, primarily used to search the binary solution space. Genetic algorithms(GAs) do not require the solution space to be continuous or need to calculate the gradient of the objective function during iteration, and as a result, the algorithm has been widely applied to solving various $\mathcal{NP}$-hard problems, such as the task scheduling problem \cite{ZXGZZ2019}, the Traveling Salesman Problem \cite{NK2012}, and the Vehicle Routing Problems \cite{Vidaletal2012}.

It is worth noting that the convergence analysis of GAs is primarily limited to convex problems or $\mathcal{P}$ problems \cite{ZH2007, QBJT2019, ZCY2021, Eibenetal1991, Rudolph1994, Rudolph1998}. However, there is a lack of theoretical results for $\mathcal{NP}$-hard problems, especially for the optimal parameter control of operators such as the mutation operator (MO)\cite{KHE2015, HM1991, LLM2007}. Despite the widespread and efficient performance of GAs in practical industrial and application processes\cite{Kramer2017}, there is still widespread skepticism among researchers regarding GA research\cite{Michalewicz2012a, Michalewicz2012b, Sorensen2015, Seretal2019}. Moreover, the No Free Lunch (NFL) theorem states that there is no superior parameterization for every problem\cite{Kramer2017, WM1997, Culberson1998, HZP2003, Koehler2007}, but it is still important to determine if there exists an optimal parameterization for a given 0-1 KP instance.

\subsection{Our results}

In this paper, we build upon the assumption of $\mathcal{NP}\neq\mathcal{P}$ \cite{Cook1971, Karp1972, GJ1979}, improve the recent theoretical results of the branch and bound algorithm (B\&B) \cite{AA1960, DDM2023}, integrating generating function approaches\cite{Brualdi2017, Jin2023}, introduce a new reduction method and an improved mutation operator (IMO). Simultaneously, the method is applicable in computing an upper bound of the mutation probability for the 0-1 KP, as well as in constructing a counterexample where the mutation probability does not tend toward 0 with an increasing number of decision variables \cite{HM1991}. The contributions of this paper can be summarized as follows.

\begin{enumerate}
\item{We propose an improved reduction method for the 0-1 KP, and provide the upper bound of leaves in the search tree (Theorem \ref{number_of_leaves}).}
\item{We propose the IMO and provide a general mathematical formulation for the upper bound of the mutation probability in solving the 0-1 KP (Theorem \ref{upper_bound}).}
\item{We construct examples where the mutation probability of the IMO does not tend towards 0 with the increase in problem size (Theorem \ref{NP!=0}).}
\item{We demonstrate that for large-scale instances, the performance of the IMO is superior to that of the MO (Theorem \ref{IMO_convergence}).}
\end{enumerate}

\subsection{Organization of remainder paper}

The remainder of this paper is organized as follows. In Section 2, we introduce the preliminary and relevant concepts of GAs. Section 3 presents the theoretical results, including a novel reduction method, the IMO proposed from this reduction method, and the associated theoretical findings. Additionally, we provide a convergence analysis of both the IMO and the MO. Finally, in the last section, we summarize the work of this paper and discuss future research directions.

\section{Preliminary and existing conclusions}

\subsection{Preliminary}

From the objective function (\ref{KP-model-obj}) and inequality (\ref{KP-model-con-1}), a primitive greedy algorithm can be easily identified: the likelihood of an item appearing in the optimal solution increases as its profit increases and its weight decreases. Consequently, we introduce the classic concept of profit density\cite{MT1990, KPP2004} to characterize the return on profit per unit weight of the items.
\begin{defi}
For $j\in N$, the profit density of item $j$ is defined by
\begin{align}
    e_j=p_j/w_j.\label{profit density}
\end{align}
\end{defi}

Without loss of generality, we always assume that the items are arranged in non-increasing order, denoted as $e_1\ge e_2\ge\dots\ge e_n$. The greedy algorithm selects items according to their profit density in descending order until the first item that cannot be packed in the knapsack. The item is known as the \emph{break item} $b$ and the solution obtained employing this greedy algorithm is the break solution. The characteristics of the break item and the break solution are defined as follows.
\begin{defi}\label{break item}
The break item $b$ is defined to be the first item that cannot be packed in the knapsack with profit density descending order, that is,
\begin{align}
    \sum\limits_{j=1}^{b-1}w_j\le C\quad\text{and} \quad \sum\limits_{j=1}^b w_j>C \label{break item-IE}
\end{align}
Correspondingly, the residual capacity $r$ is defined as
\begin{align}
    r = C-\sum\limits_{j=1}^{b-1}w_j\label{residual capacity}
\end{align}
\end{defi}
\begin{defi}\label{break solution}
The break solution is donated as $\textbf{X}^*=(x^*_1,x^*_2,\dots,x^*_n)\in\{0,1\}^n$, where $x^*_j=1$ if $j\in\{1,2,\dots,b-1\}$ and $x^*_j=0$ otherwise. For the sake of comparison, let $\textbf{Y}$ be the optimal solution.
\end{defi}

To further verify that items with higher profit density are more likely to be selected in the optimal solution, Pisinger\cite{Pisinger1995} conducted a comparison between the optimal solution $\textbf{Y}$ and the break solution $\textbf{X}^*$ for 1000 instances. Each instance had a data size of 1000, with the break item being the 500th. Pisinger found that the primary concentrations of differences between $\textbf{Y}$ and $\textbf{X}^*$ were around the break item.

Based on empirical evidence derived from the greedy algorithm and a significant number of instances, it can be empirically observed that items with higher profit density tend to be more frequently selected for inclusion in the optimal solution $\textbf{Y}$. However, it is regrettable that this observation may be challenged by a classic counterexample, which can be shown as follows.
\begin{exa}\cite{KPP2004}
Let $n=2$ and $C=M$. The first item is given by $w_1=1$ and $p_1=2$ whereas the second item is defined by $w_2=p_2=M$, where $M$ is a large constant.
\end{exa}

Clearly, the break solution is $\textbf{X}^*=(1,0)$, but the optimal solution for the counterexample is $\textbf{Y}=(0,1)$. From this counterexample, we can observe that the profit density of an item does not necessarily determine whether it will be selected in the optimal solution, and also implies that in the process of solving through GA, the optimal probability distribution for the MO cannot be determined for a given instance.

Although it is currently impossible to determine all decision variables in polynomial time complexity, with the continuous development of solving algorithms, a reduction algorithm has been discovered that can exactly solve a part of decision variables in polynomial time complexity. Given an instance $\mathscr{P}$, the reduction algorithm treats the solving process as a search tree, where each branch represents a decision variable in the problem. We employ the notation $(\mathscr{P}|x_j=\alpha)$ to denote the subproblem where the $j$-th decision variable is fixed at $\alpha\in\{0,1\}$. To facilitate subsequent discussions, $\overline{v}(\mathscr{P}|x_j=\alpha)$ is introduced as an upper bound for this subproblem. Furthermore, we introduce $\underline{v}(\mathscr{P})$ as a lower bound for the original problem $\mathscr{P}$, typically computed using the greedy algorithm or a meta-heuristic algorithm.

Despite the subproblem $(\mathscr{P}|x_j=\alpha)$ is also known to be $\mathcal{NP}$-hard, relaxing the solution space to the real field converts it into a convex problem\cite{BV2004}. Consequently, the optimal solution for the relaxed subproblem, $\overline{v}(\mathscr{P}|x_j=\alpha)$, can be obtained within polynomial time complexity. These findings lead to several significant conclusions.
\begin{thm}
     For the 0-1 KP, if $\overline{v}(\mathscr{P}|x_j=\alpha)\le \underline{v}(\mathscr{P})$, then $x_j$ is set to $1-\alpha$ in the optimal solution, where $\alpha\in\{0,1\}$.
\end{thm}

The decision variable $x_j$ is fixed, and the branch $x_j=\alpha$ is pruned. To compute $\overline{v}(\mathscr{P}|x_j=\alpha)$, the most common approach is to relax the solution space from the integer field to the real field. By doing this, we can obtain the well-known Dantzig upper bound\cite{Dantzig1957}, which is denoted as $U$ and is given by
\begin{align*}
U=\sum\limits_{j=1}^{b-1}p_j+\frac{r p_b}{w},
\end{align*}
where $r$ represents the residual capacity. It is clear that the decision variables dominated by the Dantzig upper bound $U$ can be expressed as follows:
\begin{thm}\label{reduction theorem}\cite{Pisinger1995}
    For the $j$-th item, if $e_j\ge e_b$(resp. $e_j<e_b$) and $p_j>(w_j+r)p_b/w_b$(resp. $p_j<(w_j-r)p_b/w_b$), then we have $x_j=1$(resp. $x_j=0$) in the optimal solution.
\end{thm}

The time complexity of computing the Dantzig bound is $O(n)$. It is evident that we can solve a decision variable exactly within polynomial time complexity. A significant amount of research has been conducted to improve the upper bounds obtained within the time complexity of $O(n^2)$ \cite{Pisinger1995, BZ1980, Pisinger1997, MPT1999, MPT2000}, thereby enabling the exact fixing of more decision variables within polynomial time complexity. However, in the worst-case scenario, the method cannot solve all decision variables. Consequently, for a preprocessed $\mathcal{NP}$-hard problem, to the best of our knowledge, there is currently no relevant research available regarding which decision variables are more likely to be selected in the optimal solution. Moreover, if an $\mathcal{NP}$-hard problem can be exactly solved within polynomial time complexity, heuristic algorithms, such as GAs, that can only provide locally optimal solutions within polynomial time complexity lose their significance. Based on the current state of research, we make the following assumptions.
\begin{ass}\label{NP!=P}
$\mathcal{NP}\neq\mathcal{P}$.
\end{ass}

If Assumption \ref{NP!=P} holds, then for an unpruned feasible solution, we cannot determine its status as an exact solution within polynomial time complexity. Consequently, each unpruned feasible solution has an equal probability of being the optimal solution.
\subsection{Genetic algorithm Framework}

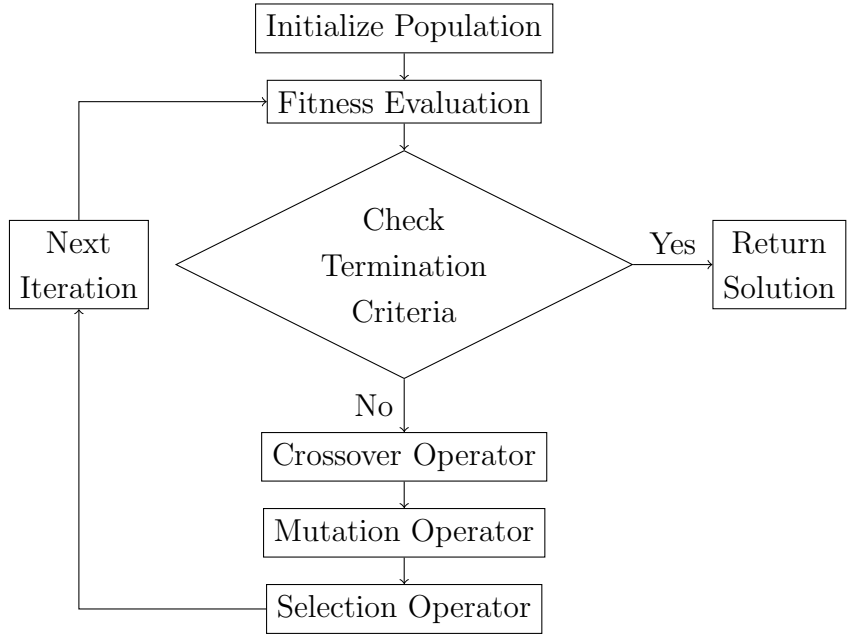
\begin{figure}[htbp]
    \centering
    \begin{tikzpicture}[node distance=10pt]
    \node[draw](1){Initialize Population};
    \node[draw, below=of 1](2){Fitness Evaluation};
    \node[draw, diamond, aspect=2, below=of 2, align=center] (3) {Check \\ Termination \\ Criteria};
    \node[draw, right=30pt of 3, align=center](4){Return \\ Solution};
    \node[draw, below=20pt of 3](5){Crossover Operator};
    \node[draw, below=of 5](6){Mutation Operator};
    \node[draw, below=of 6](7){Selection Operator};
    \node[draw, left=10pt of 3, align=center](8){Next \\ Iteration};

    \draw[->] (1) -- (2);
    \draw[->] (2) -- (3);
    \draw[->] (3) -- node[left]{No}(5);
    \draw[->] (3) -- node[above]{Yes}(4);
    \draw[->] (5) -- (6);
    \draw[->] (6) -- (7);
    \draw[->] (7) -- (8|-7) -- (8);
    \draw[->] (8) -- (8|-2) -- (2);
    \end{tikzpicture}
    \caption{Flowchart of a basic genetic algorithm
    }\label{flowchart-GA}
\end{figure}

With the prosperity of commercial activities, a vast number of business operations can be described as $\mathcal{NP}$-hard problems. Considering Assumption \ref{NP!=P}, the development of exact algorithms for solving such problems has been slow. Meanwhile, heuristic algorithms have flourished \cite{CT2018, GP2010}. As a classic algorithm in the field of heuristics, GAs have been extensively applied to solve $\mathcal{NP}$-hard problems, particularly the knapsack problem and its variants \cite{WH2017}, over the past decades. In GAs, each solution in the search space can be represented by an individual composed of several genes. By simulating natural selection, genetic mutation, and crossover operations, GAs evolve and select individuals within the current population, aiming to find the optimal solution. The flowchart of the GA can be illustrated in Figure \ref{flowchart-GA}. It is worth noting that Rudolph pointed out, through Markov chain analysis, that the GA lacks global convergence, whereas the Elitist Genetic Algorithm (EGA), which selects the current population and the optimal individual from the previous generation, possesses the property of converging to the global optimum as time approaches infinity \cite{Rudolph1994}. The main distinction between EGA and GA lies in the selection operators, while the same mutation operator is employed. Since this paper primarily focuses on the theoretical results of mutation operators, we will still adopt the representation of the GA in this study.

The GA begins by initializing the population and generating initial solutions, and the fitness value of each individual is calculated by the objective function. The algorithm then checks if the termination condition is satisfied based on the specified number of iterations. If the termination condition is met, the results are output; otherwise, the next generation population is obtained by applying the crossover, mutation, and selection operators sequentially. Common crossover operators include single-point crossover, k-point crossover, and uniform crossover. Frequent mutation operators include flip bit mutation, swap mutation, and inversion mutation. Selection operators mainly consist of roulette wheel selection, stochastic universal sampling, and rank-based selection. Various works on these operators are available in the references \cite{Schmitt2001, Mitchell1996}. In this study, we utilize the widely employed single-point crossover, flip bit mutation, and roulette wheel selection operators, along with presenting pseudo-code for the GA.

Let $O_t^k(j)$ represent the value of the $j$-th decision variable in the $k$-th individual of the $t$-th $(1\le k\le \text{pop})$ generation where $\text{pop}$ represents the population size, and let $\min f(O_t)$ (resp. $\max f(O_t)$) denote the minimum (resp. maximum) fitness value of individuals in the $t$-th generation population. Additionally, \text{LOS} refers to the local optimal solution, while $I$ and $t$ represent the maximum number of iterations and the current iteration number, respectively. Moreover, $p_m$ and $p_c$ represent the mutation probability and the crossover probability, respectively.

\begin{algorithm}[]
\caption{Genetic algorithm}\label{alg:alg1}
\begin{algorithmic}
\STATE {\textsc{Input:\ }}$n, W, P, C, \text{pop}, I, p_c, p_m$
\STATE {\textsc{Output:\ }}\text{LOS}
\STATE Generating initial population $O_1$ and let $t=1$.
\WHILE {$t\le I$}
\STATE {//Crossover Operator}
\FOR {$k=1:2:\text{pop}$}
\IF {${\rm rand}\le p_c$}
\STATE $r_t=\text{rand}*(n-1)+1, T = O_t^k$,
\STATE $O_t^k(r_t+1:n)=O_t^{k+1}(r_t+1:n),$
\STATE $O_t^{k+1}(r_t+1:n) = T(r_t+1:n).$
\ENDIF
\ENDFOR
\STATE {//Mutation Operator}
\STATE $O_{t}=\vert O_t-[{\rm rand}(\text{pop},n)\le p_m]\vert.$
\STATE {//Selection Operator}
\STATE The selection probability of the $k$-th individual is determined as $\frac{f(O_t^k)-\min f(O_t)+1} {\sum_{k'=1}^{\text{pop}} f(O_t^{k'})-\text{pop}\min f(O_t)+\text{pop}}$, and resulting in $O_{t+1}$.
\STATE $t=t+1.$
\ENDWHILE
\STATE $\text{LOS}=\max f(O_I)$
\end{algorithmic}
\label{alg1}
\end{algorithm}

The GA starts by generating initial solutions and then enters a loop. Next, the algorithm performs the crossover and mutation operators on the individuals determined by random numbers. These operations involve randomly generating crossover points, denoted as $t_r$, crossing two individuals from position $t_r$ to $n$, and flipping their bits. Afterward, the GA calculates the probability of each individual being selected using the roulette wheel selection operator. Particularly, as problem size increases, the difference between the objective function values of sub-optimal and local optimal solutions decreases, increasing the likelihood of sub-optimal solutions being selected. To address this, we assign a value of 1 to the objective function value of the worst solution in the current generation, then subtract this value from the objective function values of the other solutions, and add 1. Finally, the algorithm terminates the loop after a limited number of iterations and returns the local optimal solution (LOS).

It is worth noting that GAs were initially perceived as a robust search algorithm, leading to a lack of extensive attention to parameter control in the early stages of its development \cite{Rudolph1994}. As the research on algorithm theory progressed, however, numerous researchers acknowledged the impact of parameter control on search outcomes, and it has since become one of the most crucial issues in the field of evolutionary algorithms \cite{EHM1999}.

\section{Mainly theoretical result}
\subsection{A novel reduction method}

Recently, a new reduction method was proposed by Dey et al.\cite{DDM2023} for analyzing the performance of the B\&B in solving random integer linear programming problems. This method divides the decision variables into different regions and constrains the branching generated by decision variables within each region. 
In this section, we combine traditional reduction methods \cite{Pisinger1995} with the approach by Dey et al. \cite{DDM2023} to solve the 0-1 KP. We introduce a colored regions partitioning method for 0-1 KP, as demonstrated in Theorem \ref{Yang_reduction}. This method determines the minimum or maximum number of items that should be included in the optimal solution for each colored region.

\begin{thm}\label{Yang_reduction}
Given a set of items $N'$ where the $j$-th item satisfies the inequality
\begin{align}
\frac{p_j}{w_j+r/i}>\frac{p_b}{w_b}\label{Yang_reduction_inequality}
\end{align}
for $i\in\mathbb{N}$, we can conclude that $\sum\limits_{j\in N'}\vert x_j-y_j \vert\le i-1$, where $\textbf{Y}$ represents the optimal solution.
\end{thm}

It is evident that for the 0-1 KP, the number of leaves in the search tree for items in $N'$ is reduced from $2^{|N'|}$ to $\sum\limits_{k=1}^{i-1}\binom{|N'|}{k}$. Finding ways to effectively reduce the algorithmic complexity and improve the solving speed, particularly by exploring more efficient prune strategies between different colored regions, remains an intriguing question. Regrettably, existing research has not investigated the relationships between items within different colored regions. Therefore, we present the following definition based on Theorem \ref{Yang_reduction}.

\begin{defi}\label{divide new}
The item set $N_1=\{1,2,\dots,b-1\}$, consisting of items with a higher profit density than the break item $b$, can be divided into $m$ disjoint subsets, denoted as $N_1
=\sqcup_{i=1}^mN_{1i}$. Each $N_{1i}$ subset is defined as follows:
\begin{align}
N_{1i}=\{j|\lfloor \frac{rp_b} {p_jw_b-p_bw_j}\rfloor+1=i,e_j>e_b\}.\label{Divide-def}
\end{align}

If no item satisfies the formula (\ref{Divide-def}) for any integer $i(1\le i\le m)$, then $N_{1i}=\emptyset$.
\end{defi}
\begin{thm}\label{Divide}
Let $D_{1i}$ and $s_{1i}$ denote the item set and the number of items that are not packed into the optimal solution in $N_{1i}$, respectively, where $1\leq i\leq m$. Furthermore, we have $s_{1i}=|D_{1i}|=\sum\limits_{j\in N_{1i}}|y_j-1|$. All $s_{1i}$ form the vector $S=(s_{1i})_{i=1}^m$. Therefore, we can conclude that:
\begin{align*}
\sum\limits_{i=1}^m \frac{s_{1i}}{i} \le 1.
\end{align*}
\end{thm}
\begin{proof}
Given a vector $\textbf{Y}'$, we can determine the corresponding $D'_{1i}$ and $S'=(s'_{1i})^m_{i=1}$ of $\textbf{Y}'$ for $i=1,2,\dots,m$ using the definitions of $D_{1i}$ and $S$.

Assume, for the sake of contradiction, that $\textbf{Y}'$ is the optimal solution and satisfies
\begin{align*}
    \sum\limits_{i=1}^ms'_{1i}/i>1.
\end{align*}

It is clear that the optimal solution $\textbf{Y}'$ must not exceed the Dantzig bound. Therefore, we have
\begin{align}
\sum\limits_{i=1}^m\sum\limits_{j\in D'_{1i}} p_j\le (\sum\limits_{i=1}^m\sum\limits_{j\in D'_{1i}}w_j+r)p_b/w_b.\label{Divide-prove-1}
\end{align}

Using the formulas (\ref{Divide-def}) and Theorem \ref{Yang_reduction}, we can deduce the following inequality:
\begin{align}
\frac{\sum\limits_{i=1}^m\sum\limits_{j\in D'_{1i}}p_j}{\sum\limits_{i=1}^m\sum\limits_{j\in D'_{1i}}(w_j+r/i)}=\frac{\sum\limits_{i=1}^m\sum\limits_{j\in D'_{1i}}p_j}{\sum\limits_{i=1}^m\sum\limits_{j\in D'_{1i}}w_j+r\sum\limits_{i=1}^ms'_{1i}/i}>\frac{p_b}{w_b}\notag
\end{align}
and
\begin{align}
\sum\limits_{i=1}^m\sum\limits_{j\in D'_{1i}}p_j
& >(\sum\limits_{i=1}^m\sum\limits_{j\in D'_{1i}}w_j+r\sum\limits_{i=1}^ms'_{1i}/i)p_b/w_b
\notag\\
& >(\sum\limits_{i=1}^m\sum\limits_{j\in D'_{1i}}w_j+r)p_b/w_b.\label{Divide-prove-2}
\end{align}

Inequality (\ref{Divide-prove-1}) and inequality (\ref{Divide-prove-2}) contradict each other, proving the proposition. Hence, Theorem \ref{Divide} is established.
\end{proof}

It is worth noting that $s_{1i}$ represents the items in $N_{1i}$ that are not selected by the optimal solution, so $s_{1i}$ belongs to the set of natural numbers, i.e., $s_{1i}\in \mathbb{N}$. Even when the integer constraints of the original problem are relaxed, i.e., $x_j \in [0,1]$ and $s_{1i}\in\mathbb{R}$, the conclusion still holds. This means that
\begin{cor}\label{relation_S}
    Given a solution $\textbf{X}\in[0,1]^n$ and $s_{1i}=\sum\limits_{j\in D_{1i}}|x_j-1|$. If
    \begin{align}
        \sum\limits_{i=1}^ms_{1i}/i> 1,
    \end{align}
    then $\textbf{X}$ is not the optimal solution.
\end{cor}

Furthermore, we establish the following equivalent corollary to facilitate algorithmic computation, considering the presence of numerous empty sets $N_{1i}$ in Theorem \ref{Divide}.
\begin{cor}\label{Divide-simple-1}
Consider a vector $H=\{h_1,h_2,\dots,h_n\}$, where each element is defined as:
\begin{align*}
\begin{split}
h_j= \left \{
\begin{array}{ll}
\lfloor \frac{rp_b}{p_jw_b-p_bw_j}\rfloor+1, & \text{if}\ e_j>e_b, \\
\infty, & \text{otherwise}.
\end{array}
\right.
\end{split}
\end{align*}
If $\textbf{X}$ is the optimal solution, then the following inequality holds:
\begin{align*}
\sum\limits_{j=1}^n \frac{1-x_j}{h_j} \le 1.
\end{align*}
\end{cor}

Based on Corollary \ref{Divide-simple-1}, we can derive the following corollary for items with a profit density lower than the break item $b$.
\begin{cor}\label{Divide-simple-2}
Consider a vector $L=\{l_1,l_2,\dots,l_n\}$, where each element is defined as:
\begin{align*}
\begin{split}
l_j= \left \{
\begin{array}{ll}
\lfloor \frac{rp_b}{p_bw_j-p_jw_b}\rfloor+1, & \text{if}\ e_j<e_b, \\
\infty, & \text{otherwise}.
\end{array}
\right.
\end{split}
\end{align*}
If $\textbf{X}$ is the optimal solution, then the following inequality holds:
\begin{align*}
\sum\limits_{j=1}^n \frac{x_j}{l_j} \le 1.
\end{align*}
\end{cor}

The number of leaves pruned in the search tree of the 0-1 KP by Corollaries \ref{Divide-simple-1} and \ref{Divide-simple-2} can be expressed using generating functions\cite{Brualdi2017, Jin2023} as follows.
\begin{thm}\label{number_of_leaves}
Consider the number of leaves in the search tree, denoted as $\omega$, and the number of items in $N_{1i}$, denoted as $n_{1i}$, respectively. Additionally, let $\mathcal{X}(\lambda)$ be the polynomial representing a potential optimal solution, given by:
\begin{align}\label{number_of_leaves_equation}
\mathcal{X}(\lambda):=\prod\limits_{i=1}^m\left(\sum \limits_{j=0}^{\min \{n_{1i},i\}} {n_{1i} \choose j}\lambda^{j/i}\right).
\end{align}

Then, the number of leaves $\omega$ is equal to the sum of the coefficients of terms with an exponent no greater than 1.
\end{thm}

It is evident from Theorem \ref{number_of_leaves} that Corollary \ref{Divide-simple-1} and Corollary \ref{Divide-simple-2} significantly enhance the efficiency of B\&B algorithms in solving problems such as the 0-1 KP.

\subsection{An improvement mutation operator and the upper bound of the mutation probability}

As one of the most commonly used tools to enhance population diversity in various heuristic algorithms, the flip bit mutation operator has been widely applied due to its simplicity. Specifically, for $O_t^k(j)$, the flip bit mutation compares the mutation probability $p_m$ (typically set as $p_m \in [0.001, 0.01]$) with a randomly generated number. If the generated random number is less than $p_m$, the value of $O_t^k(j)$ is flipped, changing 1 to 0 and 0 to 1. This procedure effectively increases the diversity of the population, enabling better exploration of the search space. To facilitate subsequent discussions, we first provide the mathematical representation of the flip bit mutation.
\begin{align}\label{MO}
O_t^k(j) =
\begin{cases}
1-O_t^k(j), & \text{if } \text{rand} < p_m, \\
O_t^k(j), & \text{otherwise},
\end{cases}
\end{align}
where \text{rand} is a randomly generated number within the interval $[0,1]$.

From Theorem \ref{number_of_leaves}, it becomes evident that items positioned farther from the line formed by the break item and the origin are more likely to be selected in the optimal solution. Building upon this observation and combining it with a greedy algorithm, we propose the IMO. Given a mutation probability $p_m$, the IMO initially employs the break item to partition the set of items into two parts. Mutation operations are then performed on items belonging to different parts based on the value of $O_t^k(j)$. The mathematical representation of the IMO can be described using formula (\ref{IMO}).
\begin{figure*}[htbp]
\centering
\begin{align}\label{IMO}
O_t^k(j) =
\begin{cases}
1-O_t^k(j), & \text{if } e_j>e_b \text{ and } \text{rand} < O_t^k(j)p_m+(1-O_t^k(j))(1-p_m),\\
1-O_t^k(j), & \text{if } e_j\le e_b \text{ and } \text{rand} < (1-O_t^k(j))p_m+O_t^k(j)(1-p_m), \\
O_t^k(j), & \text{otherwise}.
\end{cases}
\end{align}
\end{figure*}

For a given mutation probability $p_m$, the IMO yields an expectation of $p_mn_{1i}$ for the number of unselected items in each set $N_{1i}$. However, due to the constraint imposed by Corollary \ref{relation_S}, there is a maximum limit on the number of unselected items among all sets. Solutions that do not meet this constraint cannot be considered optimal. Consequently, we can further reduce the ineffective search space of the mutation operator and determine an upper bound for the mutation probability $p_m$.

\begin{thm}\label{upper_bound}
Let $\overline{p}_m$ denote the upper bound of the mutation probability, which can be expressed as:
\begin{align}
    \overline{p}_m=\min\{\frac{1}{\sum_{j=1}^n 1/h_j},\frac{1}{\sum_{j=1}^n 1/l_j}\}.
\end{align}
\end{thm}
\begin{proof}
Without loss of generality, we assume that $0<\frac{1}{\sum_{j=1}^n 1/h_j}\leq \frac{1}{\sum_{j=1}^n 1/l_j}$. It is evident that, according to Corollary \ref{relation_S} and Corollary \ref{Divide-simple-1}, if $\overline{p}_m>\frac{1}{\sum_{j=1}^n 1/h_j}$, the individual processed by the IMO cannot be an optimal solution in terms of mathematical expectation. The proof of Theorem \ref{upper_bound} is thereby established.
\end{proof}

In the past decades, numerous researchers have conducted extensive research on the optimal mutation probability for convex functions or $\mathscr{P}$ problems such as \text{OneMax} and \text{LeadingOnes}. The commonly obtained value for this probability is $\frac{1}{n}$ \cite{ZH2007, QBJT2019, ZCY2021}. These studies have led to the following conclusions.

\begin{Proposition}\cite{HM1991}
With the increase in problem size, the mutation probability gradually tends to 0, i.e.,
\begin{align*}
    \lim\limits_{n\rightarrow\infty}p_m=0.
\end{align*}
\end{Proposition}

As the problem size increases, it becomes apparent that for a general instance of the 0-1 KP, the number of items with profit density greater than the break item also increases. This raises an intriguing question of whether the upper bound $\overline{p}_m$ for the mutation probability in the IMO gradually tends to 0.

\begin{thm}\label{P=0}
Let $R$ be a constant such that $R\in \mathbb{N}^+$ and consider an instance $\mathscr{P}$ of the 0-1 KP with $p_j,w_j\in \{1,2,\dots,R\}$. We can conclude that
\begin{align*}
    \lim\limits_{n\rightarrow\infty}\overline{p}_m=0.
\end{align*}
\end{thm}
\begin{proof}
    For an instance of the 0-1 KP, without loss of generality, let the profit density of the break item be constant at $p_b/w_b$, and let the residual capacity $r$ also be constant. The knapsack capacity $C$ can then be expressed as:
    \begin{align*}
        C=\sum\limits_{j=1}^{b-1}w_j+r.
    \end{align*}

    Consider the item set $N_1$ consisting of items with profit density greater than the break item $b$, and let $R$ be any constant. It is evident that there exists a constant $M\in \mathbb{N}^+$ such that $m<M$ and $N_1$ can be expressed as the disjoint union of sets $N_{1i}$, as defined in Definition \ref{divide new}. In other words, for any $j\in N_1$, we have:
    \begin{align*}
        \frac{p_j}{w_j+r/M}>\frac{p_b}{w_b}.
    \end{align*}

    Furthermore, by leveraging Corollary \ref{Divide-simple-1}, we are able to acquire a vector $H$ that is associated with the given instance. Consequently, the following conclusion can be derived:
    \begin{align*}
        \lim\limits_{n\rightarrow\infty}\overline{p}_m \le\lim\limits_{n\rightarrow\infty}\frac{1} {\sum_{j=1}^n1/h_j}
        \le\lim\limits_{n\rightarrow\infty}\frac{M} {n}=0.
    \end{align*}

    The proof of Theorem \ref{P=0} is complete.
\end{proof}

It is notable that when the profit and weight values of items in a 0-1 KP instance are limited, as exemplified in Theorem \ref{P=0}, the instance can be solved within $O(Rn^2)$ time complexity by Dynamic Programming\cite{Bellman1957, Toth1980}. Since $R$ is a constant, the instance belongs to the $\mathcal{P}$ complexity class. Consequently, it becomes an interesting question of whether the upper bound of the mutation operator in the IMO still tends towards 0 when the instance of the 0-1 KP no longer limits the profit and weight values of items.

\begin{thm}\label{NP!=0}
    For any given constant $\theta$ within the interval $(0,1)$, it is feasible to construct an instance of the 0-1 KP with $p_j,w_j\in \mathbb{N}^+$, and we have
    \begin{align*}
        \lim\limits_{n\rightarrow\infty}\overline{p}_m =\theta.
    \end{align*}
\end{thm}
\begin{proof}
    Following the same approach as the proof of Theorem \ref{P=0}, we assume that the profit density of the break item remains constant at $p_b/w_b$, while maintaining a constant residual capacity $r$. Furthermore, the knapsack capacity $C$ must satisfy
    \begin{align*}
        C=\sum\limits_{j=1}^{b-1}w_j+r.
    \end{align*}

    Without loss of generality, we set $\theta=\frac{1}{2}$. Next, we construct the profit and weight of items in the instance, ensuring that their profit densities exceed that of the break item. Let $p_j$ and $w_j$ represent the profit and weight, respectively, of the $j$-th item, satisfying
    \begin{align}\label{Divide-convergence}
        \lfloor \frac{rp_b} {p_jw_b-p_bw_j}\rfloor+1=2^{j-1}.
    \end{align}

    In other words, we have $h_j=2^{j-1}$. Since $p_j, w_j\in\mathbb{N}^+$, it is clear that there exist $p_j$ and $w_j$ that satisfy equality (\ref{Divide-convergence}). Consequently, we can conclude that
    \begin{align*}
        \lim\limits_{n\rightarrow\infty}\overline{p}_m =\lim\limits_{n\rightarrow\infty}\frac{1} {\sum_{j=1}^n1/h_j}=\frac{1}{2}=\theta.
    \end{align*}
    The proof of Theorem \ref{NP!=0} is complete.
\end{proof}

\subsection{Comparison analysis}

In $\mathcal{NP}$-hard problems, given two solutions $\textbf{X}'$ and $\textbf{X}''$, even if $\Vert \textbf{X}'-\textbf{Y} \Vert <\Vert \textbf{X}''-\textbf{Y}\Vert$, we cannot conclude that $f(\textbf{X}')\ge f(\textbf{X}'')$, where $\textbf{Y}$ is the optimal solution. In other words, solutions with a smaller Hamming distance to the optimal solution $\textbf{Y}$ do not necessarily dominate solutions with a larger Hamming distance to $\textbf{Y}$. Therefore, the existing convergence analysis of GAs is limited to specific scenarios. For a general instance of $\mathcal{NP}$-hard problems, the expected runtimes of the GA often exceed $O(2^n)$, which lacks strong theoretical guarantees for the performance.

To demonstrate the performance of the IMO, we evaluate the algorithm by considering the probability $\tau(\text{Algorithm})$ of hitting the optimal solution $\textbf{Y}$ from the zero vector $\textbf{0}$ within a single iteration for a given instance $\mathscr{P}$. This evaluation is conducted to assess the performance of the IMO and the MO on general large-scale instances.

\begin{exa}\label{example_instance}
Given an instance $\mathscr{P}$ of the 0-1 KP with $n$ items, let $\lambda_1$ and $\lambda_2$ denote the number of selected and unselected items within the first $b-1$ items, respectively. Similarly, $\lambda_3$ and $\lambda_4$ represent the number of selected and unselected items after the $(b-1)$-th item, respectively. We have
$\lambda_1 = \sum\limits_{j=1}^{b-1}|y_j - 1|$, $\lambda_2 = \sum\limits_{j=1}^{b-1}|y_j|$, $\lambda_3 = \sum\limits_{j=b}^n |y_j|$, and $\lambda_4 = \sum\limits_{j=b}^n |y_j - 1|$.
\end{exa}

\begin{thm}\label{IMO_convergence}
    If $\lambda_1<\lambda_2$ and $p_m<0.5$, then $\tau({\rm IMO})>\tau({\rm MO})$.
\end{thm}
\begin{proof}
    \begin{align}
        \frac{\tau({\rm IMO})}{\tau({\rm MO}} & =
        \frac{(1-p_m)^{\lambda_1}p_m^{\lambda_2} p_m^{\lambda_3} (1-p_m)^{\lambda_4}} {p_m^{\lambda_1+\lambda_3} (1-p_m)^{\lambda_2+\lambda_4}}\notag\\
        & = \frac{p_m^{\lambda_2+\lambda_3} (1-p_m)^{\lambda_1+\lambda_4}} {p_m^{\lambda_1+\lambda_3} (1-p_m)^{\lambda_2+\lambda_4}}\notag\\
        & = (\frac{1-p_m}{p_m})^{\lambda_2-\lambda_1}\notag\\
        &>1.\notag
    \end{align}

    This completes the proof of Theorem \ref{IMO_convergence}.
\end{proof}

Theorem \ref{IMO_convergence} demonstrates that when the number of items in the optimal solution exceeds the number of items not in the optimal solution before the break item, the probability of the IMO hitting the optimal solution in a single iteration is higher, which aligns with empirical experimental data. Moreover, since GAs are primarily employed for solving large-scale $\mathcal{NP}$-hard problems, which rarely exhibit the scenario of Example \ref{example_instance} with $\lambda_1 \geq \lambda_2$, and the mutation probability $p_m$ does not exceed $0.5$, the IMO generally outperforms the MO in large-scale instances.

\section{Conclusion and future work}

Heuristic algorithms with parameters exhibit significant performance differences depending on the value of the parameters. Due to limitations in mathematical tools, the problem of finding the optimal parameters for heuristic algorithms has received extensive attention in recent decades. Based on the assumption that $\mathcal{NP}\neq\mathcal{P}$, we propose a novel reduction method and apply it to compute the upper bound of the mutation probability.

Furthermore, our method not only proves that in the 0-1 KP, when the weight and profit of items are limited, the mutation probability tends toward 0, but also demonstrates that the mutation probability can tend toward a constant within the open interval $(0,1)$ when the weight and profit values of items are unrestricted.

For future work, we can approach it from two perspectives. Firstly, the upper bound of the method deserves further improvement. Our research only utilized the linear relaxation technique, and it would be worthwhile to explore better upper bound computation methods, such as the Lagrangian relaxation technique. Secondly, Dey et al. have demonstrated that such reduction methods can apply  to not only one-dimensional cases but also multidimensional problems. Therefore, the application scope of these reduction methods can be further expanded, such as multidimensional knapsack problems.


\begin{thebibliography}{99}
\small \setlength{\itemsep}{-.8mm}

\bibitem{MT1990} S. Martello and P. Toth, \emph{Knapsack Problems: Algorithms and Computer Implementations,} John Wiley \& Sons, Chichester, 1990.

\bibitem{KPP2004} H. Kellerer, U. Pferschy, and D. Pisinger, \emph{Knapsack Problems}, Springer-Verlag, Berlin, Heidelberg, 2004.

\bibitem{Cacchianietal2022a} V. Cacchiani, M. Iori, A. Locatelli, and S. Martello, ``Knapsack problems- An overview of recent advances. Part I: Single knapsack problems,'' \emph{Computers \& Operations Research}, vol. 143, pp. 105692, 2022.

\bibitem{Cacchianietal2022b} V. Cacchiani, M. Iori, A. Locatelli, and S. Martello, ``Knapsack problems-An overview of recent advances. Part II: Multiple, multidimensional, and quadratic knapsack problems,'' \emph{Computers \& Operations Research}, vol. 143, pp. 105693, 2022.

\bibitem{Cook1971} S. A. Cook, ``The complexity of theorem-proving procedures,'' in \emph{Proceedings of the 3rd Annual ACM Symposium on Theory of Computing(STOC)}, ACM, New York, NY, USA, 1971, pp. 151-158.

\bibitem{Karp1972} R. M. Karp, ``Reducibility among Combinatorial Problems,'' in \emph{Complexity of Computer Computations}, R. E. Miller, J. W. Thatcher, and J. D. Bohlinger, Eds. Boston, USA: Springer, 1972, pp. 85-103.

\bibitem{GJ1979} M. R. Garey and D. S. Johnson, \emph{Computer and Intractablility: A Guide to the Theory of NP-Completeness}, Freeman, San Francisco, CA, 1979.

\bibitem{Bellman1957} R. Bellman, \emph{Dynamic programming}, Princeton University Press, Princeton, 1957.

\bibitem{Toth1980} P. Toth, ``Dynamic programming algorithms for the Zero-One Knapsack Problem,'' \emph{Computing}, vol. 25, pp. 29-45, 1980.

\bibitem{HS1974} E. Horowitz and S. Sahni, ``Computing Partitions with Applications to the Knapsack Problem,'' \emph{Journal of the ACM}, vol. 21, no. 2, pp. 277-292, 1974.

\bibitem{SS1981} R. Schroeppel and A. Shamir, ``A $T = O(2^{n/2})$, $S = O(2^{n/4})$ algorithm for certain NP-complete problems,'' \emph{SIAM Journal on Computing}, vol. 10, no. 3, pp. 456-464, 1981.

\bibitem{SBR2002} M. V. Sapir, J.-C. Birget, and E. Rips, ``Isoperimetric and isodiametric functions of groups,'' \emph{Annals of Mathematics}, vol. 156, no. 2, pp. 345-466, 2002.

\bibitem{DS2005} I. Dinur and S. Safra, ``On the hardness of approximating vertex cover,'' \emph{Annals of Mathematics}, vol. 162, no. 1, pp. 439-485, 2005.

\bibitem{Holland1975} J. H. Holland, \emph{Adaptation in Natural and Artificial Systems}, MIT Press, 1975.

\bibitem{ZXGZZ2019} H. Zhang, J. Xie, J. Ge, Z. Zhang, and B. Zong, ``A hybrid adaptively genetic algorithm for task scheduling problem in the phased array radar,'' \emph{European Journal of Operational Research}, vol. 272, no. 3, pp. 868-878, 2019.

\bibitem{NK2012} Y. Nagata and S. Kobayashi, ``A Powerful Genetic Algorithm Using Edge Assembly Crossover for the Traveling Salesman Problem,'' \emph{INFORMS Journal on Computing}, vol. 25, no. 2, pp. 346-363, 2012.

\bibitem{Vidaletal2012} T. Vidal, T. G. Crainic, M. Gendreau, N. Lahrichi, and W. Rei, ``A Hybrid Genetic Algorithm for Multidepot and Periodic Vehicle Routing Problems,'' \emph{Operations Research}, vol. 60, no. 3, pp. 611-624, 2012.

\bibitem{ZH2007} Y. Zhou and J. He, ``A Runtime Analysis of Evolutionary Algorithms for Constrained Optimization Problems,'' \emph{IEEE Transactions on Evolutionary Computation}, vol. 11, no. 5, pp. 608-619, 2007.

\bibitem{QBJT2019} C. Qian, C. Bian, W. Jiang, and K. Tang, ``Running Time Analysis of the (1 + 1)-EA for OneMax and LeadingOnes Under Bit-Wise Noise,'' \emph{Algorithmica}, vol. 81, pp. 749-795, 2019.

\bibitem{ZCY2021} W. Zheng, H. Chen, and X. Yao, ``Analysis of Evolutionary Algorithms on Fitness Function with Time-linkage Property,'' \emph{IEEE Transactions on Evolutionary Computation}, vol. 25, no. 4, pp. 696-709, 2021.

\bibitem{Eibenetal1991} A. E. Eiben, E. H. L. Aarts, and K. M. Van Hee, ``Global convergence of genetic algorithms: A markov chain analysis,'' in \emph{Parallel Problem Solving from Nature}. PPSN 1990, H.-P. Schwefel and R. Manner, Eds. Springer, Berlin, Heidelberg, 1991, pp. 507-516.

\bibitem{Rudolph1994} G. Rudolph, ``Convergence analysis of canonical genetic algorithms,'' \emph{IEEE Transactions on Neural Networks}, vol. 5, no. 1, pp. 96-101, 1994.

\bibitem{Rudolph1998} G. Rudolph, ``Finite Markov Chain Results in Evolutionary Computation: A Tour d'Horizon,'' \emph{Fundamenta Informaticae}, vol. 35, no. 1-4, pp. 67-89, 1998.

\bibitem{KHE2015} G. Karafotias, M. Hoogendoorn, and A. E. Eiben, ``Parameter Control in Evolutionary Algorithms: Trends and Challenges,'' \emph{IEEE Transactions on Evolutionary Computation}, vol. 19, no. 2, pp. 167-187, 2015.

\bibitem{HM1991} J. Hesser and R. Manner, ``Towards an optimal mutation probability for genetic algorithms,'' in \emph{Proceedings of the 1st Conference on Parallel Problem Solving from Nature}, Springer, Dortmund, Germany, 1991, pp. 23-32.

\bibitem{LLM2007} F. G. Lobo, C. F. Lima, and Z. Michalewicz, \emph{Parameter Setting in Evolutionary Algorithms}, Springer-Verlag Berlin Heidelberg, 2007.

\bibitem{Kramer2017} O. Kramer, \emph{Genetic Algorithm Essentials}, Springer, 2017.

\bibitem{Michalewicz2012a} Z. Michalewicz, ``Ubiquity symposium: Evolutionary computation and the processes of life: the emperor is naked: evolutionary algorithms for real-world applications,'' \emph{Ubiquity}, vol. 2012, no. November, pp. 1-13.

\bibitem{Michalewicz2012b} Z. Michalewicz, ``Quo Vadis, Evolutionary Computation?'' in \emph{Advances in Computational Intelligence}, J. Liu, C. Alippi, B. Bouchon-Meunier, G. W. Greenwood, and H. A. Abbass, Eds., vol. 7311, Springer, Berlin, Heidelberg, 2012.

\bibitem{Sorensen2015} K. Sorensen, ``Metaheuristics-the metaphor exposed,'' \emph{International Transactions in Operational Research}, vol. 22, pp. 3-18, 2015.


\bibitem{WM1997} D. H. Wolpert and W. G. Macready, ``No free lunch theorems for optimization,'' \emph{IEEE Transactions on Evolutionary Computation}, vol. 1, no. 1, pp. 67-82, Apr. 1997.

\bibitem{Culberson1998} J. C. Culberson, ``On the futility of blind search: An algorithmic view of 'No Free Lunch','' \emph{Evolutionary Computation}, vol. 6, no. 2, pp. 109-127, Jun. 1998.

\bibitem{HZP2003} Y. C. Ho, Q. C. Zhao, and D. L. Pepyne, ``The No Free Lunch Theorems: Complexity and Security,'' \emph{IEEE Transactions on Automatic Control}, vol. 48, no. 5, pp. 783-795, May 2003.

\bibitem{Koehler2007} G. J. Koehler, ``Conditions that Obviate the No-Free-Lunch Theorems for Optimization,'' \emph{INFORMS Journal on Computing}, vol. 19, no. 2, pp. 273-279, May 2007.

\bibitem{AA1960} A. H. Land and A. G. Doig, ``An automatic method of solving discrete programming problem,'' \emph{Econometrica}, vol. 28, no. 3, pp. 497-520, Jul. 1960.

\bibitem{DDM2023} S. S. Dey, Y. Dubey, and M. Molinaro, ``Branch-and-bound solves random binary IPs in poly($n$)-time,'' \emph{Mathematical Programming}, vol. 200, pp. 569-587, 2023. Accepted for publication in SODA 2021.

\bibitem{Brualdi2017} R. A. Brualdi, \emph{Introductory Combinatorics}, Pearson, Fifth Edition, 2017.

\bibitem{Jin2023} E. Y. Jin, ``Symmetric generating functions and Euler-Stirling statistics on permutations,'' \emph{Journal of Combinatorial Theory, Series A}, vol. 197, p. 105752, Mar. 2023.

\bibitem{Pisinger1995} D. Pisinger, ``An expanding-core algorithm for the exact 0-1 Knapsack Problem,'' \emph{European Journal of Operational Research}, vol. 87, no. 1, pp. 175-187, Nov. 1995.

\bibitem{BV2004} S. Boyd and L. Vandenberghe, \emph{Convex Optimization}, Cambridge University Press, 2004.

\bibitem{Dantzig1957} G. B. Dantzig, ``Discrete-Variable Extremum Problems,'' \emph{Operations Research}, vol. 5, no. 2, pp. 266-288, Apr. 1957.

\bibitem{BZ1980} E. Balas and E. Zemel, ``An Algorithm for the Large Zero-One Knapsack Problems,'' \emph{Operations Research}, vol. 28, no. 5, pp. 1130-1154, Sep. 1980.

\bibitem{Pisinger1997} D. Pisinger, ``A Minimal Algorithm for the 0-1 Knapsack Problem,'' \emph{Operations Research}, vol. 46, no. 5, pp. 758-767, Oct. 1997.

\bibitem{MPT1999} S. Martello, D. Pisinger, and P. Toth, ``Dynamic programming and strong bounds for the 0-1 knapsack problem,'' \emph{Management Science}, vol. 45, no. 3, pp. 414-424, Mar. 1999.

\bibitem{MPT2000} S. Martello, D. Pisinger, and P. Toth, ``New trends in exact algorithms for the 0-1 knapsack problem,'' \emph{European Journal of Operational Research}, vol. 123, no. 2, pp. 325-332, Jun. 2000.

\bibitem{WH2017} X. Wang and Y. He, ``Evolutionary Algorithms for Knapsack Problems,'' \emph{Journal of Software}, vol. 28, no. 1, pp. 1-16, Jan. 2017. (in Chinese)

\bibitem{Schmitt2001} L. M. Schmitt, ``Theory of genetic algorithms,'' \emph{Theoretical Computer Science}, vol. 259, no. 1-2, pp. 1-61, Nov. 2001.

\bibitem{Mitchell1996} M. Mitchell, \emph{An Introduction to Genetic Algorithms}, MIT Press, 1996.

\bibitem{EHM1999} A. E. Eiben, R. Hinterding, and Z. Michalewicz, ``Parameter Control in Evolutionary Algorithms,'' \emph{IEEE Transactions on Evolutionary Computation}, vol. 3, no. 2, pp. 124-141, Jul. 1999.

\bibitem{CT2018} B. Chopard and M. Tomassini, \emph{An Introduction to Metaheuristics for Optimization}, Springer, 2018.

\bibitem{GP2010} M. Gendreau and J. Y. Potvin, \emph{Handbook of Metaheuristics}, Springer, 2nd Edition, 2010.

\end{thebibliography}
\end{document}